\documentclass{article}

\usepackage{arxiv}

\usepackage[utf8]{inputenc} 
\usepackage[T1]{fontenc}    
\usepackage{hyperref}       
\usepackage{url}            
\usepackage{booktabs}       
\usepackage{amsfonts}       
\usepackage{nicefrac}       
\usepackage{microtype}      
\usepackage{lipsum}		
\usepackage{graphicx}
\usepackage[numbers]{natbib}
\usepackage{doi}
\usepackage{algorithm}
\usepackage{algorithmic}
\usepackage{xcolor}
\usepackage{microtype}
\usepackage{graphicx}
\usepackage{subfigure}
\usepackage{booktabs} 
\usepackage{float}
\usepackage{stfloats}

\usepackage{amsmath}
\usepackage{amssymb}
\usepackage{mathtools}
\usepackage{amsthm}

\DeclareMathOperator*{\argmin}{argmin}
\DeclareMathOperator*{\argmax}{argmax}
\theoremstyle{plain}

\theoremstyle{definition}

\theoremstyle{remark}

\title{Towards Robust Deep Active Learning for Scientific Computing}
\setcitestyle{square}

\author{ Simiao Ren, Yang Deng, Willie J. Padilla and Jordan M. Malof\\
	Department of Electrical and Computer Engineering\\
	Duke University\\
	Durham, NC 27705, USA \\
	\texttt{jordan.malof@duke.edu} \\
}

\date{}

\renewcommand{\shorttitle}{Towards Robust Deep Active Learning for Scientific Computing}

\hypersetup{
pdftitle={Hyperparameter-free deep active learning for regression problems via query synthesis},
pdfauthor={Simiao Ren, Yang Deng, Willie J. Padilla, Jordan Malof},
pdfkeywords={Deep Learning , Active Learning , Regression , Query-by-committee , Query Synthesis , Inverse Problem , Scientific Computing , Artificial Electromagnetic Material},
}

\begin{document}
\maketitle

\begin{abstract}
    Deep learning (DL) is revolutionizing the scientific computing community. To reduce the data gap, active learning has been identified as a promising solution for DL in the scientific computing community. However, the deep active learning (DAL) literature is dominated by image classification problems and pool-based methods. Here we investigate the robustness of pool-based DAL methods for scientific computing problems (dominated by regression) where DNNs are increasingly used.  We show that modern pool-based DAL methods all share an untunable hyperparameter, termed the pool ratio, denoted $\gamma$, which is often assumed to be known apriori in the literature.  We evaluate the performance of five state-of-the-art DAL methods on six benchmark problems if we assume $\gamma$ is \textit{not} known - a more realistic assumption for scientific computing problems. Our results indicate that this reduces the performance of modern DAL methods and that they sometimes can even perform worse than random sampling, creating significant uncertainty when used in real-world settings.  To overcome this limitation we propose, to our knowledge, the first query synthesis DAL method for regression, termed NA-QBC. NA-QBC removes the sensitive $\gamma$ hyperparameter and we find that, on average, it outperforms the other DAL methods on our benchmark problems. Crucially, NA-QBC always outperforms random sampling, providing more robust performance benefits. 
\end{abstract}


\section{Introduction}
\label{sec:introduction}

Deep learning has led to major advances in many areas of scientific computing \cite{jumper2021highly,rolnick2019tackling, lavin2021simulation,khatib2021deep}, however one of its major limitations is the need for large quantities of labeled data.  One widely-studied method to reduce the data needs of machine learning models is active learning (AL) \cite{ren2021survey, settles2009active}.  A large number of active learning methods have been developed for deep neural networks (DNNs) - called deep active learning (DAL) methods \cite{roy2018deep}.  In this work we investigate DAL for \textit{regression problems}, which has become increasingly important in recent years due to emerging applications of DNNs in many areas of science: e.g., chemistry \cite{schutt2019unifying}, materials science \cite{nadell2019deep}, and biology \cite{zhavoronkov2019deep} where DNNs are employed to model the properties of natural systems. 

There are three well-known active learning paradigms \cite{settles2009active} : (a) pool-based (b) stream-based (c) and query synthesis (QS).  We focus here on \textit{pool-based} methods since they currently dominate the literature of active learning \cite{ren2021survey, kumar2020active}.  In pool-based DAL, the model must choose the best instances to label from a finite set of unlabeled candidates, termed \textit{the pool}.  One important hyperparameter common to all pool-based methods is the \textit{pool ratio}, denoted $\gamma = N_{U} / K$, which is the ratio of the pool's size, $N_{U}$, to the number of points $K$ that we aim to select from the pool for labeling in DAL each iteration.  Previous work has indicated that active learning models can perform poorly if $\gamma$ is set improperly. For example, an excessively large $\gamma$ value can cause a condition known as mode collapse  \cite{burbidge2007active, ren2021survey, kee2018query} (see Sec. "problem setting" for further details).  In the literature, $\gamma$ values are often assumed to be known apriori, although to our knowledge there is no general method for \textit{optimizing} $\gamma$ without trial-and-error \textit{using labeled data, and therefore defeating the purpose of active learning}.   

In this work we examine the robustness of current DAL methods on scientific computing problems, namely what happens if we relax the assumption that $\gamma$ is \textit{not} assumed to be known apriori, as would be the case in \textit{all} real-world applications.  To do this, we conduct two experiments.  First, we evaluate and compare the performance of five DAL regression methods on six different regression problems, as we vary their $\gamma$ parameter. Our benchmark problems include four contemporary problems from science and engineering.  Second, we propose a cross-validation experiment in which we optimize $\gamma$ on one problem, assuming access to all of its labels, and then apply it to the remaining datasets, reflecting a real-world strategy for choosing $\gamma$. To our knowledge this is the first benchmark comparison of modern DAL methods for scientific computing (regression), and the first analysis of their sensitivity to $\gamma$. 

Our results indicate that the performance of modern DAL methods often varies significantly with respect to $\gamma$, and more importantly, that the best setting of $\gamma$ varies widely across problems. This is consistent with the widely-varying values of $\gamma$ we find in the literature (see Sec "problem setting").  Furthermore, some settings of $\gamma$ also yield performance that is worse than simple random sampling.  These results suggest a substantial uncertainty when deploying DAL to a new problems, where $\gamma$ is generally unknown. We assert that this greatly undermines the value of DAL for real-world applications.

To overcome this limitation we propose - to our knowledge - the first QS DAL method for regression (see Sec. related work).  Our approach, termed NA-QBC, relies upon the widely-used query-by-committee (QBC) criteria. We frame DAL as an inverse problem and then employ the recently-proposed neural-adjoint optimizer to efficiently search the input space with gradient ascent for instances that maximize the QBC measure.  Crucially, the resulting NA-QBC method no longer requires a pool, and therefore avoids the $\gamma$ hyperparameter.  On our benchmark experiments we find that NA-QBC achieves performance comparable to our baseline methods if they are applied with their optimized $\gamma$ settings.  Furthermore, in our cross-validation experiments we find that NA-QBC achieves the best average performance across our benchmarks, and crucially, is always superior to random sampling. 

\subsection{Contribution of this work}

\begin{itemize}
    
    \item The first public benchmark for pool-based DAL scientific computing regression problems and analysis of sensitivity to their pool ratio.
    
    \item NA-QBC: the first query synthesis deep active learning model for scientific computing regression problems, which eliminates the $\gamma$ hyperparameter, a huge step towards robust and usable DAL for scientific computing community.
    
\end{itemize}

\section{Related works}\label{sec:related_work}


\textbf{Query synthesis active learning} Although QS was proposed relatively early on \cite{angluin1988queries}, it has yet to make the same impact as pool-based methods, due to its difficulty to work together with human annotators \cite{baum1992query}. Interpolation techniques between labeled data points were first investigated by \cite{baum1991neural} for the synthesized query. Following interpolation, \cite{wang2015active} proposed synthesizing the middle point of the closest opposite pair, which is restricted to classification problems. \cite{king2004functional, king2009automation} employed QS in real-life settings for automating biological experiments and achieved a much lower cost than humans. \cite{cohn1996active} derived the statistical optimal choice of a new query point that minimizes the variance of the learner but pointed out too many approximations were needed for neural networks to effectively use such techniques. \cite{englhardt2020exploring} explored QS on one-class classifiers with evolutionary algorithms, non-trivial to convert to a regression setting. With deep learning, \cite{zhu2017generative, mahapatra2018efficient, mayer2020adversarial} used Generative Adversarial Networks (GAN) to synthesize new images for classification tasks, however transforming the GAN techniques into regression tasks is not intuitive due to the inherent "discriminator" portion of GAN. Although the GAN portion is hardly transferable to our problem setting, the gradient descent steps to search the space is close to NA-QBC in that both method use gradient information to guide the search for new query points.

The lack of attention for QS methods can also be seen in recent review papers, two recent reviews on active learning, while both acknowledging the existence of QS paradigm, did not discuss any specific QS algorithm \cite{kumar2020active, ren2021survey}. To our knowledge, only one review paper discussed QS \cite{settles2009active} howerver, as it was published in the pre-ImageNet era, none of the QS strategies mentioned involved or has a simple pathway to be adapted to deep learning.
 
\textbf{Active learning for regression problems } Although significant effort has pushed the boundary of active learning, only a few studies have focused on regression tasks. For example, in an ICML workshop on active learning and experimental design, the "Active Learning Challenge" \cite{guyon2011results} had all of their 6 benchmark datasets being binary classification tasks. For regression tasks, Expected model change \cite{settles2008curious} was explored, where an ensemble of models was used \cite{cai2013maximizing} to estimate the true label of new query point. Gaussian Process \cite{kading2018active} were used with a natural estimate of variance on unlabeled points. \cite{smith2018less} used QBC, which trains multiple networks and finds the most disagreeing unlabeled points of the committee of models trained. \cite{tsymbalov2018dropout} used the Monte Carlo drop-out under a Bayesian setting, also aiming for the maximally disagreed points. Although these studies have investigated regression tasks, none of the above used QS paradigm.

\section{Problem Setting} 
Our problem setting of scientific computing problems is fundamentally different from the majority of the pool-based active learning literature, which mainly focuses on image classification problems, in two ways:
\begin{enumerate}
    \item Scientific computing mainly focuses on regression problems instead of classification problems
    \item Images live on a high dimensional 'valid' manifold, as only a small set of 2D pixel intensity maps are actually natural images. On the other hand, scientific computing usually has a clearly defined input range, usually, a hyper-cube defined by the practitioner or physics law, and all input within the hyper-cube is a 'valid' input.
\end{enumerate}
\label{sec:problem_setting}

Our problem setting is formally defined as: Let $T^i = (X^{i}, Y^{i})$ be the dataset used to train a regression model at the $i^{th}$ iteration of active learning.  We assume access to some oracle (e.g., a simulator for scientific computing problems, or human annotator for image classificatioin problem), denoted $f : \mathcal{X} \rightarrow \mathcal{Y}$, that can accurately produce the target values, $y \in \mathcal{Y}$ associated to input values $x \in \mathcal{X}$.  Since our focus is on DAL, we assume a DNN as our regression model, denoted $\hat{f}$.  We assume that some relatively small number of $N_{0}$ labeled training instances are available to initially train $\hat{f}$, denoted $T^0$.  In each iteration of DAL, we must get $K$ query instances $x_{k} \in \mathcal{X}$ to be labeled by the oracle, yielding a set of labeled instances, denoted $L$, that is added to the training dataset. Our goal is then to choose $L$ that maximize the performance of the DNN-based regression models over  unseen test data at each iteration of active learning.  

\textbf{Pool-based Deep Active Learning.} General pool-based DAL methods assume that we have some pool $U$ of $N_{U}$ unlabeled instances from which we can choose the $K$ instances to label. Most pool-based methods rely upon some function $q: \mathcal{X} \rightarrow \mathbb{R}$ to assign some scalar value to each $x \in U$ indicating its "informativeness", or utility for training $\hat{f}$. In each iteration of active learning, $q$, is used to evaluate all instances in $U$, and the top $K$ are chosen to be labeled and included in $T$.  This general algorithm is outlined in Algorithm \ref{alg:QBC}.

\begin{algorithm}[h]
  \caption{Generic pool-based active learning algorithm}
  \label{alg:QBC}
    \begin{algorithmic}
      \STATE {\bfseries Input:} Initial labeled training set $T^{0}$ of size $N_{ini}$, step size $K$, AL criteria function $q$, number of steps $I$
      \FOR{$i=0$ {\bfseries to} $I$}
          \STATE Train DNN-based model(s) using training set $T^{i}$
          \STATE Create $U$ by sampling $N_{U}$ instances $x \in \mathcal{X}$ 
          \STATE Calculate  $q(x) \; \forall x \in U$
          \STATE Create $L$ by labeling top $K$ points in $U$ ranked by $q(x)$
          \STATE $T^{i+1} = T^{i} \cup L$
      \ENDFOR
    \end{algorithmic}
\end{algorithm}

An important distinction between pool-based DAL approaches application scenarios and our setting is the ability to sample, or synthesize, new values of $x$.  In many problems (e.g., natural imagery or audio data) the data live on some manifold in a high-dimensional space that is unknown or otherwise difficult to sample \cite{sener2017active,wang2015active,roy2018deep,ash2019deep}. In these settings $U$ typically consists of some predefined and fixed set of unlabeled instances that must be used throughout active learning.   

By contrast, in many scientific computing problems $\mathcal{X}$ can be readily sampled and it is possible to change $U$ throughout the process of active learning and data subspace is defined apriori by a designer, or otherwise constrained by physical laws.  For example, all of our benchmark problems satisfy this criterion.  Due to the use of DNNs in science and engineering this setting is increasingly important, and it is our focus in this work.  In this setting, pool-based DAL methods often regenerate/resample $U$ after each iteration \cite{smith2018less}, which we adopt here in Algorithm \ref{alg:QBC}.   


\begin{figure}[h]
    \begin{center}
    \centerline{\includegraphics[width=\linewidth]{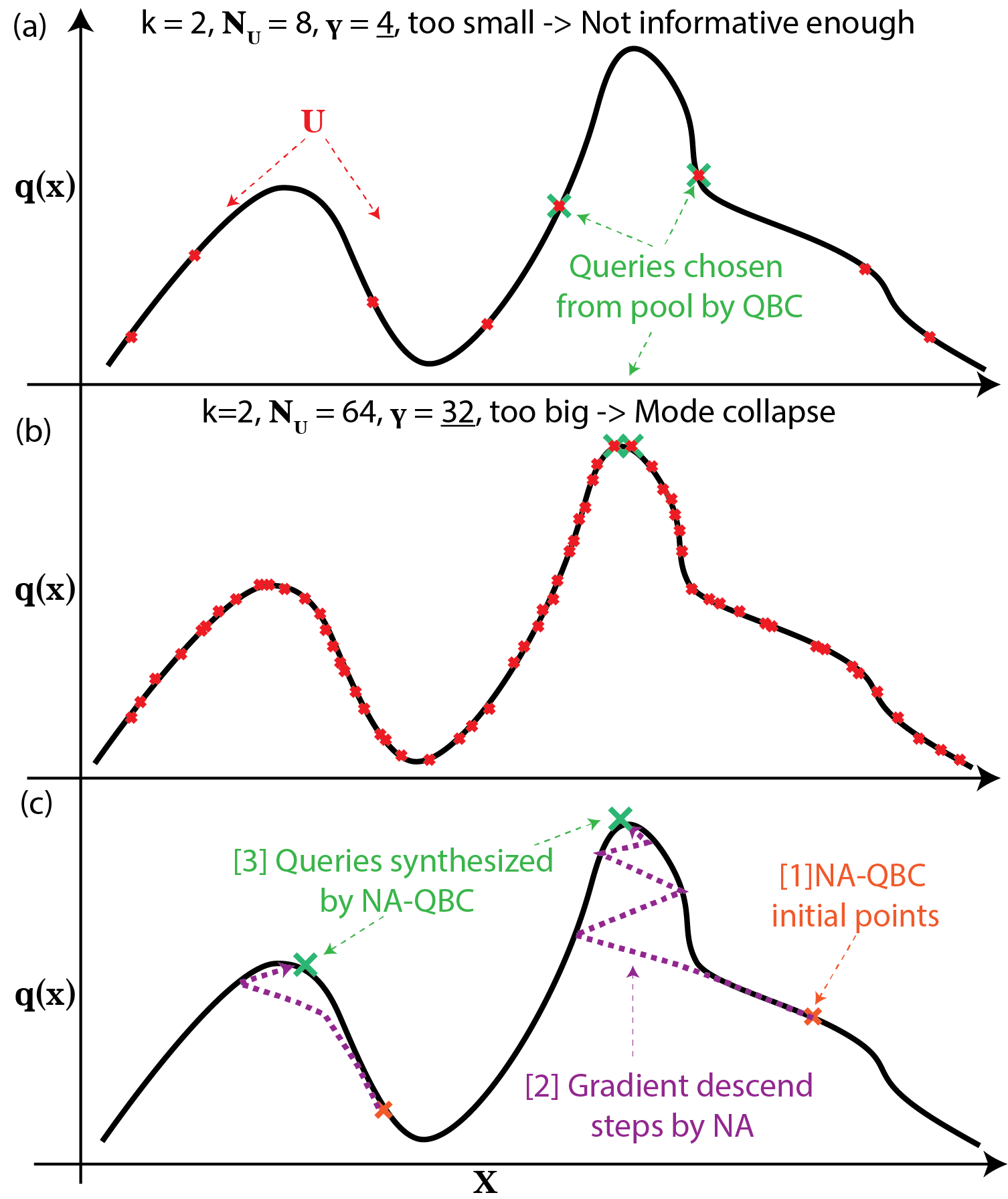}}
    \caption{Schematic diagram for pool-based DAL and NA-QBC mechanism. $q(x)$ is the acquisition metric. (a, b) are two scenarios of the pool ratio ($\gamma$) being too small (4 in b) or too large (32 in c) in $K$ (step size) of 2. (c) Working mechanism of NA-QBC. }
    \label{img:QBC_illustration}
    \end{center}
\end{figure}

\textbf{The pool ratio hyperparameter.} One consequence of our problem setting is that we can use any pool size, $N_{U}$.  A larger value of $N_{U}$ can lead to the discovery of points with larger values of $q(x)$ because the input space is sampled more densely; however, larger $N_{U}$ also tends to increase the similarity of the points, so that they provide the same information to the model - a problem sometimes called mode collapse \cite{burbidge2007active, ren2021survey, kee2018query}. In the limit as $N_{U} \rightarrow \infty$ all of the $k$ selected query points will be located in near the same $x \in \mathcal{X}$ that has the highest value of $q(x)$. The negative impact of excessively small and large values of $N_{U}$, respectively, is illustrated in Fig. \ref{img:QBC_illustration}(a-b) for a simple 1-dimensional problem.  The step size, $k$, of the active learning model also interacts strongly with the pool size, and therefore the ratio of pool size to step size is treated as a single hyperparameter,
\begin{equation}
    \gamma = N_{U}/k . 
\end{equation}
Crucially, and as we show in our experiments, choosing a sub-optimal $\gamma$ value can result in poorer performance than naive random sampling. This isn't necessarily a problem if either (i) one $\gamma$ setting works across most problems or, alternatively, (ii) $\gamma$ can be optimized on new problems without using labels.  We are not aware of any method for optimizing $\gamma$ on a new problem without first collecting large quantities of labels.  Furthermore, the value of $\gamma$ varies widely across the literature, suggesting that suitable settings for $\gamma$ indeed vary across problems: e.g., 17000 \cite{mccallumzy1998employing}, 20 to 2000 \cite{kee2018query}, 300 to 375\cite{santos2020modeling}, 11-20 \cite{roy2018deep}, 1000 \cite{burbidge2007active}, and 1 to 11 \cite{tan2019batch}. We also corroborate these findings on five state-of-the-art DAL methods on our six benchmark problems in Sec "result".  This sensitivity to $\gamma$ greatly undermine the value of modern DAL methods since their benefits can vary, and they may even be inferior to simple random sampling.



\section{Neural adjoint QBC} \label{sec:NA-QBC}

We begin this section by first introducing the query-by-committee (QBC) method, which is the basis of several state-of-the-art DAL methods, as well as NA-QBC.  QBC is a well-known pool-based active learning algorithm \cite{seung1992query} where the goal is to choose the top $K$ samples from $U$ as determined by the following measure, known as the QBC criterion:
\begin{equation}
    \label{eq:query_by_committee_criterion}
    q_{QBC}(x) = \frac{1}{N}\sum^N_{n=1}(\hat{f}_n(x)-\mu(x))^2
\end{equation}
Here $\hat{f}_{n}$ denotes the $n^{th}$ model in an ensemble of $N_{ens}$ models (DNNs in our case), and $\mu(x)$ is the mean of the ensemble predictions at $x$. In each iteration of AL these models are trained on all available training data at that iteration.  Substituting $q_{QBC}$ for the function $q$ in Algorithm \ref{alg:QBC} results in the QBC DAL method, which will form one of the baselines for our experiments.  

\textbf{Neural-adjoint QBC.} The NA-QBC method attempts to \textit{efficiently} search over all $x \in \mathcal{X}$ to find values that maximize the $q_{QBC}$, which can be framed as the following optimization problem 
\begin{equation}
    \label{eq:na_qbc_optimization_problem}
    \argmax_{x \in \mathcal{X}}  q_{QBC}(x) \\
\end{equation}

\begin{algorithm}[H]
  \caption{Neural-adjoint Query-by-committee (NA-QBC) query synthesis active learning}
  \label{alg:NAQBC}
    \begin{algorithmic}
      \STATE {\bfseries Input:} Initial labeled training set $T_0$, step size $k$, number of active learning steps $S$, models $f_i$, number of backpropagation steps $S_{bp}$, learning rate $\alpha$ 
      \FOR{$s=1$ {\bfseries to} $S$ }
          \FOR{$n=1$ {\bfseries to} $N$}
            \STATE Train $f_n$ using training set $T$
          \ENDFOR
          \STATE randomly initialize $k$ points in $x$ space as $X_{add}$
          \FOR{$x_j \in X_{add}$}
            \FOR{$s_{bp}=1$ {\bfseries to} $S_{bp}$}
                \STATE Calculate $\mathcal{L} =$ $\mathcal{L}_{boundary} - \sigma^2$
                \STATE Update: $x_j \mathrel{-}= \alpha \dfrac{d\mathcal{L}}{dx_j}$
            \ENDFOR
          \ENDFOR
          \STATE Let oracle label the $X_{add}$, add them into $T$ 
      \ENDFOR
    \end{algorithmic}
\end{algorithm}

The solutions to this optimization then are synthesized and labeled.  By assumption in Sec. "problem setting", $\mathcal{X}$ is a well-defined space of admissible solutions and that we have an oracle that can accurately label any $x \in \mathcal{X}$, which are reasonable assumptions for a wide array of problems. 

To solve this optimization problem we leverage the recently-proposed Neural Adjoint (NA) method \cite{ren2020benchmarking}, which was shown to efficiently minimize (or maximize) an arbitrary black-box function, denoted $q(x)$ here, with respect to its input $x$. A major assumption of NA is that $q$ can be accurately approximated with a DNN, $\hat{q}$, making it possible to efficiently minimize $\hat{q}$ with respect to its input $x$ using gradient descent (via backpropagation), starting from some randomly-initialized location within the domain of $x$.   More formally, let $\hat{x}^{i}$ be our current estimate of the solution, where $i$ indexes each solution we obtain in an iterative gradient-based estimation procedure.  Then we compute $\hat{x}^{i+1}$ with  
\begin{equation}    
\hat{x}^{i+1} = \hat{x}^{i} + \alpha  \left. \frac{\partial  (\hat{q}_{QBC}(x) -  \mathcal{L}_{bnd}(x))}{\partial x} \right\rvert_{x=\hat{x}^{i}} \label{eq:nagrad} 
\end{equation}
where $\alpha$ is the learning rate, which can be made adaptive like Adam \citep{kingma2014adam}. The term $\mathcal{L}_{bnd}$ is a loss term that prevents the gradient-based search from escaping the the space of valid solutions, $\mathcal{X}$. The boundary loss is defined by
\begin{equation} \label{eq:Lbdy}
    \mathcal{L}_{bnd}(x) = \lambda * 
    \begin{cases}
        x - x_{max}, & \text{if } x \geq x_{max}\\
        0, & \text{if } x_{min} \leq x \leq x_{max}\\
        x_{min} - x, & \text{if } x \leq x_{min}
    \end{cases}
\end{equation}

where $\lambda$ is the boundary strength, $x_{max, min}$ are the dimension wise extrema. This is implemented as $\lambda * ReLU(|x - x_{mid}| - 0.5R_x) $ when the range of x is in a hyper-cube. ReLU is rectified linear unit, $x_{mid}$ is the middle point of the hypercube of our problem domain and $R_x$ is the dimension-wise range. The hyperparameter $\lambda$ controls the cost for leaving $\mathcal{X}$.  

The original NA \cite{ren2020benchmarking} initializes Eq. \ref{eq:nagrad} with a large number of randomly-sampled $x$-values in an effort to find a globally optimal solution. The best solution among the candidates is then found by passing each converged solution back into $\hat{q}$ and retaining the best one.  We adapt NA for the DAL problem by setting $N_{ini} = K$, the number of query points we wish to label, and then retaining all resulting solutions after convergence. Another difference of our application of NA is that since the committee is trained during each AL step, we did not train an extra "proxy" model for variance but used the committee directly as our forward model in NA. The final NA-QBC algorithm is outlined in Algorithm \ref{alg:NAQBC}.

\textbf{The NA-QBC hyperparameters} Although NA-QBC no longer has the $\gamma$ hyperparameter, the use of the neural-adjoint optimizer does introduce two extra hyperparameters: Boundary loss strength and learning rate (during synthesis), which we describe further in the Supplement.  Importantly though, and in contrast to $\gamma$ however, we find that both parameters are (i) insensitive to changes in the learning problem; and (ii) that they can be adjusted at any point of DAL, without the need for any additional labeled data (i.e., they can be tuned prior to applying DAL) as we can monitor the synthesized x at any given point.  To show to robustness of our algorithm with this hyper-parameter, we fixed the boundary loss strength $\lambda$ as 1 for all experiments in results.

\section{Benchmark Regression Problems} 
\label{sec:benchmark_problems}

We propose six regression problems to include in our public DAL regression benchmark, four of which represent contemporary problems from diverse fields of science and engineering. Although relatively low-dimensional (e.g., compared to natural imagery or audio), several of these problems are still complex and required expressive DNN-based models in their original publication.  Furthermore, the evaluation of the Oracle in most of these problems is time-consuming and/or costly, making data collection a major bottleneck. The remaining two problems are simpler, and were included primarily to support model analysis. Major properties of each problem are listed in Table \ref{tbl:benchmark_dataset}.  One other selection criterion was the availability of an oracle function, which is needed to label novel query points, such as those found by NA-QBC.  We briefly describe each problem below, but further technical details can be found in the supplement.  Upon publication, we will publish all benchmark resources to support future study.

\textbf{1D sine wave (SINE).} A noiseless 1-dimensional sinusoid with smoothly-varying frequency.  

\textbf{2D robotic arm (ARM)} \cite{ren2020benchmarking} In this problem we aim to predict the 2-D spatial location of the endpoint of a robotic arm based upon its three joint angles, $x$.  

\textbf{Stacked material (STACK)} \cite{Chen2019} The goal is to predict the 201-D reflection spectrum of a material based upon the thickness of five layers of the material. 

\textbf{Artificial Dielectric Material (ADM)} \cite{deng2021neural} The goal is to predict the 2000-D reflection spectrum of a material based upon based upon its 14-D geometric structure. Full wave electromagnetic simulations were utilized in \cite{deng2021benchmarking} to label data, requiring 1-2 minutes per input instance. 

\textbf{NASA Airfoil (FOIL)} \cite{Dua:2019} The goal is to predict the sound pressure of an airfoil based upon structural properties of the foil, such as its angle of attack and chord length.  This problem was recently published by NASA\cite{brooks1989airfoil}. and the instance labels were obtained from a series of real-world aerodynamic tests in an anechoic wind tunnel.  

\textbf{Hydrodynamics (HYDR)} \cite{Dua:2019} The goal is to predict the residual resistance of a yacht hull in water based upon its shape.  This problem was recently published by the Technical University of Delft (hosted by UCI ML repository \cite{Dua:2019}), and the instance labels were obtained by real-world experiments using a model yacht hull in water. 

\section{Baseline Active Learning Methods}
\label{sec:baseline_active_learning_methods}

We include five baseline in our benchmarks, to our knowledge comprising all existing active learning methods in the literature that could be applied to (i) deep neural networks (ii) for regression problems, \textit{without requiring significant modification}.  We briefly describe each method below, and refer readers to other references for full details.  Upon publication, we will publish software for all of these methods to support future benchmarking. 

\textbf{Query-by-committee (QBC)} \cite{seung1992query} The QBC approach is described in Sec. "problem setting", and given by Alg. \ref{alg:QBC} if we set $q(x) = q_{QBC}(x)$.  It is also illustrated in Fig. \ref{img:QBC_illustration}.

\textbf{QBC with diversity (Div-QBC)} \cite{kee2018query}  This method improves upon QBC by adding a term to $q$ that also encourages the selected query points to be diverse from one another. This method introduces a hyperparameter for the relative weight of the diversity and QBC criteria and we use an equal weighting, as done in the original paper \cite{kee2018query}. 
    
\textbf{QBC with diversity and density (DenDiv-QBC)} \cite{kee2018query}  This builds upon Div-QBC by adding a term to $q(x)$ that encourages query points to have uniform density. This method introduces two new hyperparameters for the relative weight of the density, diversity, and QBC criteria, and we use an equal weighting as done in the original paper \cite{kee2018query}. 

\textbf{BALD (MC Dropout)} \cite{tsymbalov2018dropout} BALD use Monte Carlo dropout technique to produce multiple probabilistic model output to estimate the uncertainty of model output and use that as the criteria of selection. Empirically we found the MC dropout model harder to fit the regression task compared to our ensemble, therefore we trained two separate network (one for AL, one for regression) to ensure fairness of evaluation. 

\textbf{Core-set} \cite{sener2017active} Unlike our other benchmarks, this approach only relies upon the density of points in the input space, $\mathcal{X}$, when selecting new query locations. A greedy selection criteria is used, given by $\max_{i\in U} \min_{j\in L'} \Delta(x_i, x_j)$ where $L'$ represents the training set plus the points already selected as queries to be labeled in the current step.

\section{Benchmark Experiment Design} \label{sec:exp}

\begin{table}[h]
    \caption{Set up for our benchmark experiments. $Dim_{x, y}$ are the dimensionality of x and y. Number of committee (\# COM) is fixed at 10 except for ADM problem where GPU RAM limits it to 5.}
    \label{tbl:benchmark_dataset}
    \begin{center}
        \begin{small}
        \begin{sc}
            \begin{tabular}{lcccccc}
            \toprule
            Feat & Sine & Robo & Stack  & ADM & Foil & Hydr\\
            \midrule
            $Dim_{x}$&      1   &   4   &   5   &   14 & 5 & 6\\
            $Dim_{y}$&      1   &   2   &   201 &   2000 & 1 & 1\\
            $e^{*}$ &  1e-3    &   5e-5    &   3e-5    &   3e-3 & 3e-3 & 7e-3    \\
            \# com &  10      &   10      &   10      &   5     & 10 & 10  \\
            $N_{0}$&      \multicolumn{6}{c}{80}    \\
            $K$&          \multicolumn{6}{c}{40} \\
            $N_{test}$&  \multicolumn{6}{c}{4000} \\
            \bottomrule
            \end{tabular}
        \end{sc}
        \end{small}
    \end{center}
\end{table}

In our experiments we will compare NA-QBC to five state-of-the-art DAL methods for regression on six different scientific computing problems.  We have two primary objectives: (i) evaluate the sensitivity of these methods to the pool ratio, $\gamma$; and (ii) evaluate their performance under real-world conditions, where we cannot optimize $\gamma$ with labeled data on each problem before running DAL.  We conduct one experiment to address each of these two objectives. Before describing the experimental design details, we first describe our proposed DAL performance measure. 

\textbf{Normalized annotation burden ($\eta$).} The most widely-used performance metric is the accuracy of $\hat{f}$, given a specific quantity of training data, denoted $N_{T}$ \cite{kee2018query, guyon2011results, tsymbalov2018dropout, cohn1996active, mccallumzy1998employing}. This is useful for many tasks where the total quantity of data is a major practical constraint, and we wish to minimize regression error given some fixed value of $N_{T}$. In scientific computing however, we often have some target level of error that we want our regression model to reach, denoted $e$, and we wish to minimize the quantity of data that must be labeled - usually via costly hand annotation, or time-consuming simulation - to achieve $e$, or lower.  Therefore we propose to measure error in terms of "annotation burden", given by 
\begin{equation} \label{eq:eta_unnormalized}
\tilde{\eta}_{m}(e) = \argmin_{N_{T}} e(N_{T},m) \leq e,
\end{equation}
where $e(N_{T},m)$ is the error of $\hat{f}$ given some quantity of training data, $N_{T}$, which was sampled using DAL method $m$. The value of $\tilde{\eta}_{m}$ can vary widely across problems, and therefore we normalize it by $\tilde{\eta}_{rand}$. We call the resulting metric "Normalized Annotation Burden", given by
\begin{equation} \label{eq:eta_normalized}
    \eta_{m}(e) = \tilde{\eta}_{m}(e) / \tilde{\eta}_{rand}(e).
\end{equation}
Therefore, if $\eta_{m}(e)=0.8$ it means that DAL method $m$ only required 80$\%$ of the data required by random sampling to reach the same level of error, $e$.  Generally, if $\eta_{m}(e) < 1.0$ it implies that DAL method $m$ was more effective than random sampling, which we consider essential for any useful DAL method. To our knowledge, a comparable performance metric has never been utilized specifically for active learning.   We use mean squared error for $e$ when computing $\eta$ for all of our benchmark regression experiments.  

\textbf{Exp. 1: pool ratio sensitivity.}  We evaluate the performance of our DAL methods as a function of $\gamma$ on each of our benchmark problems, with $\gamma \in \Gamma = \{2^k\}_{k=1}^{6}$ . Following convention \cite{kee2018query,tsymbalov2018dropout}, we assume a small training dataset is available at the outset active learning, $T_{0}$, which has $N_{0}$ randomly sampled training instances. We then run each DAL model until the regression model, $\hat{f}$, reaches some desired level of mean-squared-error, denoted $e^{*}$, as evaluated empirically on some withheld testing dataset that is shared by all DAL methods.  Because the empirically-measured test error is noisy, we stop active learning only when $e \leq e^{*}$ for five steps, and use the fifth step to compute $\eta_{m}(e^{*})$.  For each benchmark problem, we assume an achievable error level, $e^{*}$ and appropriate neural network architecture is known apriori. We obtain these settings from previously reported benchmarks (e.g., the Stack and ADM problems), or through experimentation.  The target value of $e^{*}$ and the size of the network are reported in Table \ref{tbl:benchmark_dataset} for each task.   Each experiment (i.e., combination of dataset, DAL model, and $\gamma$ value) is run 5 times.  For each DAL model, $m$, we compute 25 values of $\eta_{m}(e)$ corresponding to each possible pair of the 5 random sampling models, and the 5 models with DAL method $m$. 

\textbf{Exp. 2: Real-world performance estimation.} To estimate the real-world performance of DAL methods, we propose a cross-validation experiment where we optimize $\gamma$ on just one of our benchmark problems, assuming access to all of its labeled data, and then use the optimized value for another problem where it is assumed we do not have access to labeled data, and therefore cannot optimize $\gamma$.  As discussed in Sec. "introduction", we are not aware of any general method to optimize $\gamma$ on some new problem without first collecting labels, defeating the purpose of DAL.  Let $\eta_{m}(e,i \rightarrow j)$ be the performance obtained on dataset $j$ using the $\gamma$ value optimized on dataset $i$, where $i,j\in{1,..,6}$.  Then the final performance we report is given by
\begin{equation} \label{eq:cross_validation}
\eta^{cv}_{m}(e) = \dfrac{1}{n}\sum_{i} \eta_{m}(e,i \rightarrow j)
\end{equation}
which is the average performance on dataset $j$ when using pool ratios found from all other datasets. This measure can be viewed as the performance on dataset $j$ when sampling from the distribution of $\gamma$ values one would obtain from real-world experimentation.  

\section{Experimental Results} \label{sec:result}
\begin{figure*}[h]
    \begin{center}
    \centerline{\includegraphics[width=\linewidth]{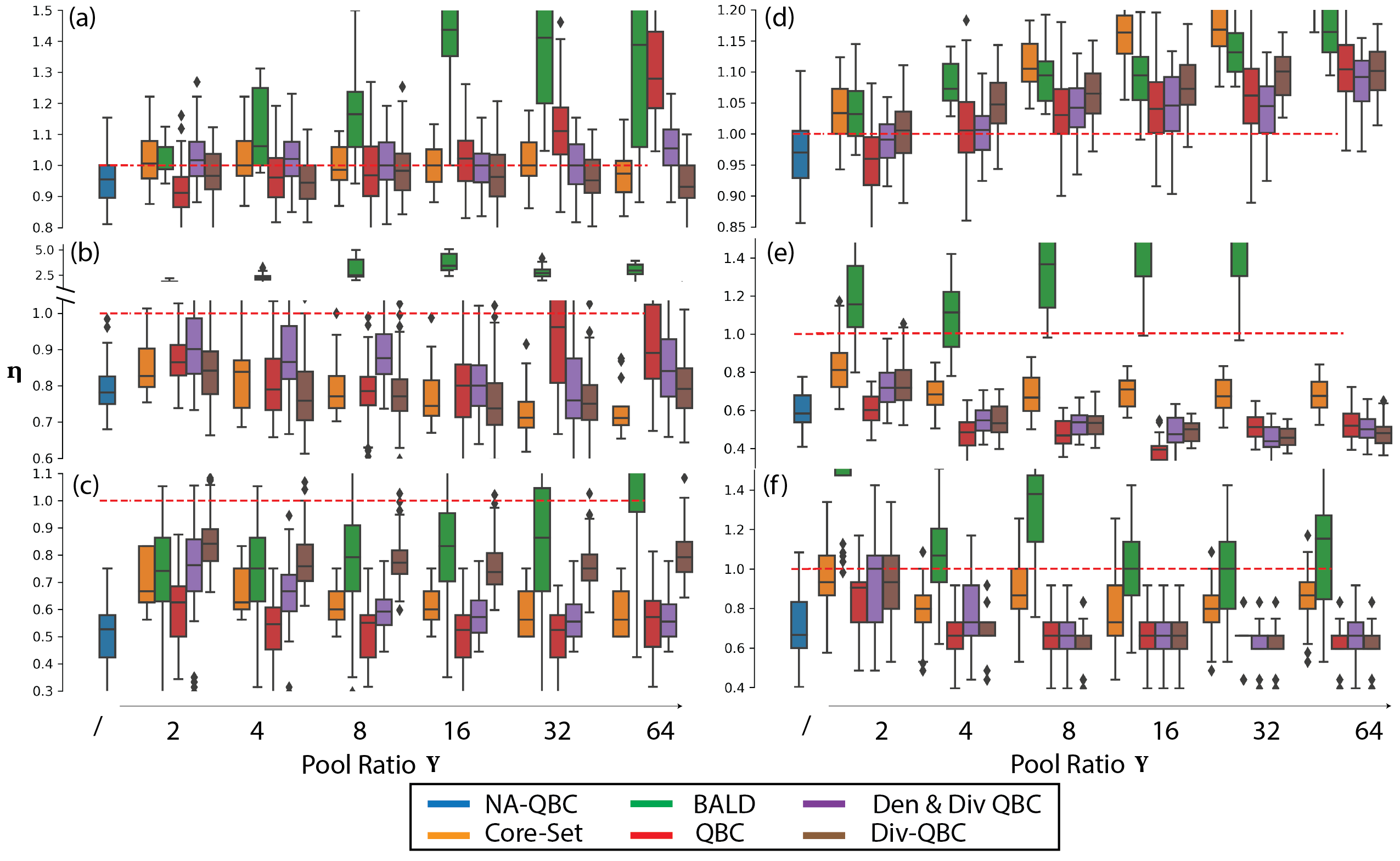}}
    \vskip -0.1in
    \caption{Normalized Annotation Burden ($\eta$) for each of our six DAL methods as a function of pool ratio ($\gamma$), on each of our six benchmark tasks. Note that for NA-QBC, it does not have a candidate pool therefore the pool ratio is labeled as "/".  The red dashed line corresponds to performance of random sampling $\eta=1$. The datasets are (a) SINE, (b) ROBO, (c) STACK, (d) ADM, (E) FOIL, and (F) HYDR. }
    \label{img:main_perf}
    \end{center}
\end{figure*}

Fig \ref{img:main_perf} presents the results of Experiment 1, and indicate a strong sensitivity to $\gamma$ for all pool-based methods.  The performance a particular DAL method on a given task often varies significantly with respect to $\gamma$, but not in a consistent way. As a consequence, there is no setting of $\gamma$ that works well across all tasks for a particular DAL method.  Furthermore, some settings of $\gamma$ also lead to performance that is worse than simple random sampling (i.e., $\eta>1$, or above the red line), and the setting that leads to this outcome varies widely as well. For example, on the ADM problem (d) a low setting of $\gamma=2$ yields the best performance for most methods, but the worst performance for most of the same methods on Foil and Hydro datasets (e,f).

In summary, due to their sensitivity to $\gamma$ it is unclear whether a given DAL method will perform any better than random sampling on a new task, and there is a risk that it may even perform worse. This uncertainty, paired with their additional complexity, dramatically undermines the value of DAL for real-world applications and consistent performance advantages is crucial.  We hypothesize that the sensitivity of DAL methods to $\gamma$, and other hyperparameters, leads many scientific computing researchers to avoid their use.  This sensitivity is frequently ignored in the literature, e.g., where it is often assumed that good hyperparameters are known.  We suggest that building more robust DAL methods is an important open and under-studied area of research.   
\begin{figure}[h]
    \begin{center}
    \centerline{\includegraphics[width=\linewidth]{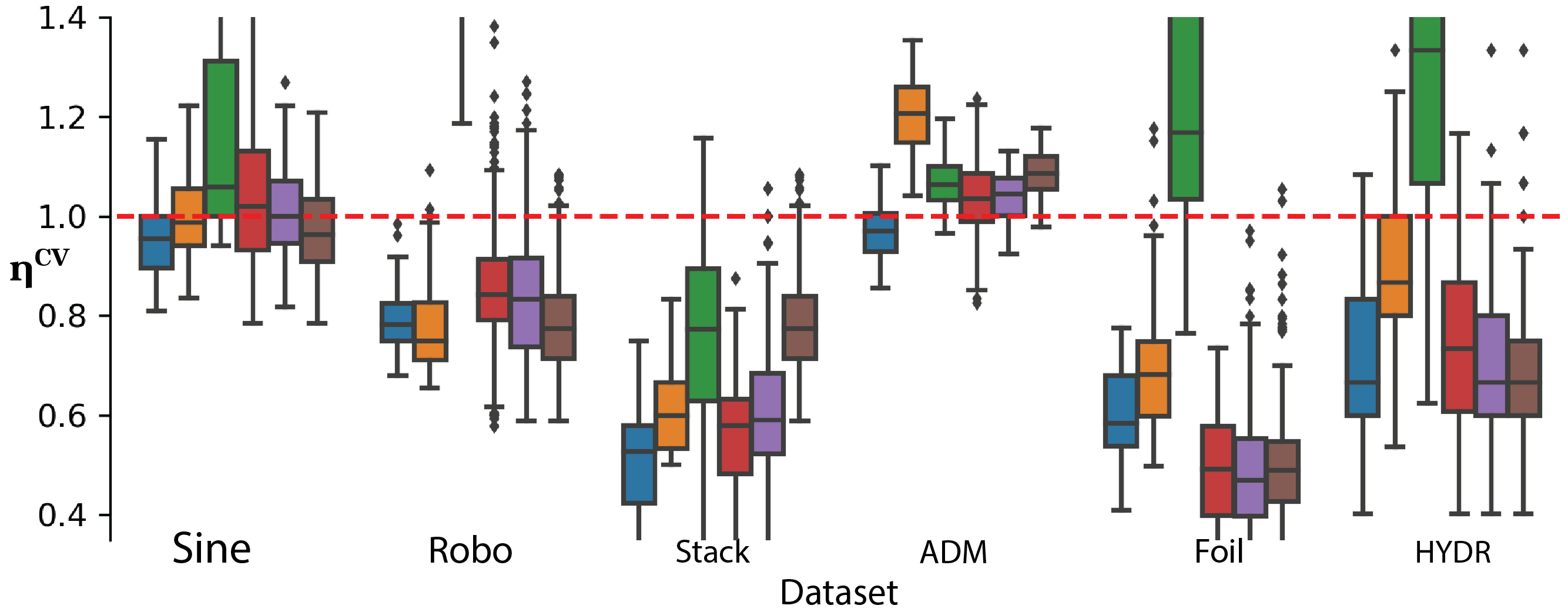}}
    \caption{Real world performance $\eta_{m}^{cv}$ comparison across all datasets. As defined in Eq \ref{eq:cross_validation}, the optimal pool ratio $\gamma$ is chosen from the mean data burden at the $e$ of each of the rest datasets and the $\eta_{m}^{cv}$ is combined to a single box}
    \label{img:round_robin}
    \end{center}
\end{figure}

Towards addressing this limitation, we proposed NA-QBC, which removes the $\gamma$ hyperparameter.  We see in Fig. \ref{img:main_perf} that, although NA-QBC does not always perform best, it never performs worse (on average) than random sampling.  Furthermore, it achieves performance similar to other methods when they have optimal $\gamma$ settings. In Experiment 2 we account for the uncertainty of $\gamma$ by optimizing it on one dataset (i.e.,assuming all labels are available), and then using this optimized setting on all other datasets.  The results of this experiment are presented in Fig. \ref{img:round_robin} where we see that NA-QBC often achieves the best performance, and achieves the best average performance across all benchmark tasks. Furthermore, it always performs better than random sampling on average, while other methods perform similar or worse to random sampling on two of our six problems (Sine and ADM).

\section{Conclusion} \label{sec:conclusions}
In this work, we present the hyperparameter selection issue faced by the popular active learning pool-based method when applying to scientific computing problems in six benchmark regression problems. We find that the highly sensitive pool ratio $\gamma$ makes QBC much limited in application to new regression problems and, by considering the query selection process as an inverse problem and combining QBC with inverse solver Neural Adjoint, we innovate NA-QBC as the first query synthesis deep active learning algorithm that is robust as it does not depend on any pool to select queries for next step, making it an effectively zero hyperparameter version of QBC approach. We also propose a data burden metric $\eta$ that better captures the effect of active learning in scientific computing scenarios and shows NA-QBC's robustness on different problems using this metric. Finally, we test our zero hyperparameter version NA-QBC on real-world performance with cross-dataset generalization metric  $\eta_{real}$ where $\gamma$ is optimized in a known dataset and tested on a new dataset. The cross-dataset experiments show NA-QBC being a competitive AL method compared to 5 different baselines when considering the hyperparameter selection process.

\bibliographystyle{ieeetr}
\bibliography{references}  
\end{document}